\documentclass{article}
\usepackage{spconf}
\usepackage{cite}
\usepackage{stfloats}
\usepackage{graphicx, subfigure}
\usepackage{float}
\usepackage{amsmath}
\usepackage{amssymb,amsthm,bm}
\usepackage{enumerate}
\usepackage{epstopdf}
\usepackage{url}
\usepackage{svg}

\title{Multi-period Time Series Modeling with Sparsity via Bayesian Variational Inference}
\name{Daniel Hsu}
\address{Department of Electrical and Computer Engineering\\
Georgia Institute of Technology\\
Atlanta, GA, 30332}
\begin{document}
\maketitle

\begin{abstract}
In this paper, we use augmented the hierarchical latent variable model to model multi-period time series, where the dynamics of time series are governed by factors or trends in multiple periods. Previous methods based on stacked recurrent neural network (RNN) and deep belief network (DBN) models cannot model the tendencies in multiple periods, and no models for sequential data pay special attention to redundant input variables which have no or even negative impact on prediction and modeling. Applying hierarchical latent variable model with multiple transition periods, our proposed algorithm can capture dependencies in different temporal resolutions. Introducing Bayesian neural network with Horseshoe prior as input network, we can discard the redundant input variables in the optimization process, concurrently with the learning of other parts of the model. Based on experiments with both synthetic and real-world data, we show that the proposed method significantly improves the modeling and prediction performance on multi-period time series.
\end{abstract}

\section{Introduction}
Time series forecasting and modeling is an important interdisciplinary field of research, involving among Computer Sciences, Statistics, and Econometrics. Made popular by Box and Jenkins \cite{bj} in the 1970s, traditional modeling procedures combine linear autoregression (AR) and moving average. But, since data are nowadays abundantly available, often complex patterns that are not linear can be extracted. So, the need for nonlinear forecasting procedures arises. Moreover, many sequential data in practice are influenced by factors in multiple periods, and control variables we collect may not be all relevant with target data. For example, sales data have daily, weekly and monthly effects, and some control variables, such as oil price, currency exchange rate and sales of some other products, may not be relevant. 

Recently, neural networks with deep architectures have proven to be very successful in image, video, audio and language leaning tasks \cite{nature}.  In time series forecasting area, though traditionally shallow neural networks are generally adopted, the deep neural networks have also aroused enormous interests among researchers. Deep belief networks (DBN) are frequently employed in current short-term traffic forecasting \cite{huang}\cite{lv}, and pre-training strategies with unsupervised learning algorithms such as Restricted Boltzmann machine (RBM) \cite{rbm} and Stacked AutoEncoder (SAE) \cite{sae} are also used. However, these deep architectures can not capture long dependencies across data points which are beyond the observation window. 

RNNs are particularly suitable for modeling dynamical systems as they operate on input information as well as a trace of previously acquired information (due to recurrent connections) allowing for direct processing of temporal dependencies. RNNs can be employed for a wide range of tasks as they inherit their flexibility from plain neural networks. Among all RNN architectures, the most successful ones to characterize long-term memory are the long short-term memory network (LSTM) \cite{lstm} and Gated Recurrent Unit (GRU) \cite{gru}, which learn both short-term and long-term memory by enforcing constant error flow through the designed cell state. However, these models still have some disadvantages. Especially LSTM model cannot work well on cases where the prediction is primarily based on recent past observations \cite{first}. Recently some papers have applied LSTM or GRU to model nonlinear sequential data by discovering latent variables, such as Variational Recurrent Neural Network (VRNN) \cite{vrnn} and Stochastic Recurrent Neural Network (SRNN) \cite{srnn}, which have achieved successful results in modeling time series with complex dynamics. Although some work such as Recurrent Ladder Network \cite{rln} applied multi-layer latent variables to model complex time series, the dynamics of all latent variables is one-step, which cannot capture transitions of time series in multiple periods. And input data collected in practice may have redundant variables, which have little or even negative effects on time series modeling. None of classical models above can deal with this issue. 

In this work, we investigate time series modeling by introducing multiple latent variables with different transition steps, and discard redundant input variables by Bayesian sparse learning with noncentered Horseshoe prior \cite{bs}. In this model, the input observations are first processed by a sparsified neural network, which is to discard irrelavent input variables. Then in the inference network composed by stacked GRUs, we can estimate the latent variables in all layers with different transition steps. The transition of latent variable in each layer is modeled by mutli-variate normal distribution conditioned on the processed input data, the latent variables in previous time step and the latent variable in last layer. The decoder is a multi-layer neural network which uses all current latent variables to predict the target time series. The model is learned by Bayesian variational inference \cite{vi1}\cite{vi2}.

The experiments show that this model can not only improve the modeling and prediction performance on both synthesis data and real-world data. The synthesis data is generated by a state-space model with latent variables in multiple layers, and each layer has different transition step. The real-world data comes from the Rossmann sales data in Germany. It is publicly available on Kaggle platform, and consists of daily sales records in 1115 branches, ranging from January 1, 2013 to July 31, 2015.

\section{Preliminary}

\subsection{Gated Recurrent Unit}
The LSTM neural network is adopted in this study to model time series. In order to resolve the vanishing gradient problem of RNN, LSTM was initially introduced in \cite{lstm}, which can model long-term dependencies and capture the temporal correlation at different time scales. Recently, as a variant of LSTM, Gated Recurrent Unit (GRU) was proposed in \cite{gru}. Compared with LSTM, it has simpler structure and competitive performance. Similarly to the LSTM unit, the GRU has gating units that modulate the flow of information inside the unit, however, without having a separate memory cells. The operations in the GRU cell are described as below,
\begin{equation}
\label{state}
\begin{split}
&\bm{r}_t=\sigma(\bm{W}_r\bm{x}_t+\bm{U}_r\bm{h}_{t-1}+\bm{b}_r) \\
&\bm{u}_t=\sigma(\bm{W}_u\bm{x}_t+\bm{U}_u\bm{h}_{t-1}+\bm{b}_u) \\
&\bm{c}_t=\sigma(\bm{W}_c\bm{x}_t+\bm{U}_c(\bm{r}_t\odot\bm{h}_{t-1})+\bm{b}_c) \\
&\bm{h}_t=\bm{u}_t\odot\bm{h}_{t-1}+(1-\bm{u}_t)\odot\bm{c}_t 
\end{split}
\end{equation}
where $\bm{x}_t$ is the input, $\bm{h}_t$ is the hidden state (activation) of GRU cell, $\bm{r}_t$ is the reset gate, $\bm{u}_t$ is the forgetting gate, and $\bm{c}_t$ is the candidate activation. $\bm{W}_{\cdot}, \bm{U}_{\cdot}$ and $\bm{b}_{\cdot}$ are weight matrices and bias in state transition, which need to be learned in training. Here $\odot$ is the elementwise multiplication. 

The hidden state $\bm{h}_t$ of GRU is the linear interpolation between the previous hidden state and candidate activation $\bm{c}_t$, where a forgetting gate $\bm{u}_t$ controls how much the unit forgets its previous hidden state. This procedure of taking a linear sum between the existing state and the newly computed state is similar to the LSTM unit. The GRU, however, does not have any mechanism to control the degree to which its state is exposed, but exposes the whole state each time. The reset gate $\bm{r}_t$ controls the influence of previous hidden state on candidate activation. In this paper, we adopt GRU as the basic transition unit for each latent variable.

\subsection{Bayesian Neural Network with Horseshoe Prior}
For certain neural network, such as the input network in our model, we denote $\omega_{kl}\in\mathbb{R}^{K_{l-1}+1\times 1}$ as all weights attached on neuron $k$ of hidden layer $l$. Authors in \cite{horseshoe} introduced sparsity-inducing prior such that the weight vector of each unit is conditionally independent and follow a group Horseshoe prior \cite{hs},
\begin{eqnarray}
\omega_{kl}|\tau_{kl}, \nu_{kl}&\sim &\mathcal{N}(0, (\tau^2_{kl}\nu^2_{kl})\mathbb{I}) \nonumber \\
\tau_{kl}\sim C^+(0, b_0), && \nu_l\sim C^+(0, b_g) \label{hs_prior}
\end{eqnarray}
where $\mathbb{I}$ is the identity matrix and $x\sim C^+(0, a)$ is the Half-Cauchy distribution for $x>0$. Here $\tau_{kl}$ is the neuron specific scale parameter, controlling the sparsity of weights associated with neuron $k$ at layer $l$, while the scale parameter $\nu_{l}$ is shared across the layer, controlling the overall sparsity of layer $l$. 

Although the horseshoe achieve some successful achievement in sparsifying neural networks, the correlations between the weights $\omega_{kl}$ and scales $\tau_{kl}\nu_{kl}$ give rise to coupled posteriors exhibiting pathological funnel shaped geometries \cite{hs2} which are difficult to sample. The non-centered parameterizations proposed in \cite{hs2} can alleviate this problem. We can reformulate the weights as below,
\begin{equation}
\omega_{kl}=\tau_{kl}\nu_{l}\beta_{kl}, \hspace{15pt} \beta_{kl}\sim\mathcal{N}(0, \mathbb{I}) \label{hs_prior2}
\end{equation}
Such a parameterization is referred to as non-centered, since the scales and weights are sampled from independent prior distributions and are marginally uncorrelated. The likelihood is now responsible for introducing the coupling between the two, when conditioning on observed data. 
In experiments \cite{horseshoe} the non-centered parameterization can significantly improve the quality of the posterior approximation for BNNs with Horseshoe priors. Thus we adopt non-centered parameterization with Horseshoe prior here. 

\section{Related Work}
Driven by the recent success of deep learning \cite{nature}, several different deep learning approaches can be found in the literature for performing time series predictions. For example, deep belief networks are used in the work of \cite{rbm} along with restricted Boltzman machine (RBM). \cite{comp} also compares the performance of Deep Belief Networks with that of Stacked Denoising Autoencoders. This last type of network is also employed by \cite{sae} to predict the temperature of an indoor environment. Another method for time series forecasting can be found in \cite{lv}, which uses Stacked Autoencoders (SAE) to predict the flow of traffic from a big data dataset. However, as compared in \cite{thesis}, deep learning models such as RBM and SAE perform worse than LSTM because they cannot capture long-term dependencies across data points, and fixed-size input window also leads to sub-optimal performance.

LSTM and GRU \cite{lstm} is another learning structure often used in time series prediction. \cite{first} first used LSTM to predict chaotic time series. In \cite{gmm}, an LSTM sequence-to-sequence model was used to predict next values. A survey \cite{fmb} reviews many applications of LSTM to short-term load forecasting problem. However, the sequence-to-sequence RNN models \cite{gmm} can not handle very long sequences or model periods very well in time series prediction \cite{ernn}. Every input sequence has to be padded to the same length. As indicated in \cite{first}, LSTM model such as \cite{xma} cannot utilize recent observations effectively, since it spends too much resources on long-term dependencies. 

%The time series forecasting with multi-output and multi-step is an active research topic. \cite{svr} investigated this problem by employing multiple-output support vector regression (MSVR) with multiple-input multiple-output (MIMO) prediction strategy. In \cite{cmtl}, a cooperative neural evolution method for multi-task learning was adopted, which enables neural networks to be trained with shared knowledge representation.

\section{Multi-period Deep Markov Model}
In this work, we adopt multi-period deep Markov model (MP-DMM), which is a hierarchical state space model with different transition periods, to improve the modeling performance of multiple temporal dependencies in sequential data. Here, at time step $t$, we use $\bm{x}_t, \bm{z}^l_t, \bm{h}^l_t$ and $\bm{y}_t$ to denote the input variables, latent variable at layer $l$, hidden state of GRU at layer $l$, and target data respectively. In each layer $l$, the transition period is denoted as $S_l$, where $S_1=1$ and $S_{l-1}\le S_l$, for $l\le L$. A sparse neural network $\varphi(\cdot)$ is used to filter out irrelevant or negative input variables for modeling and prediction. The dataset $\mathcal{D}$ contains all input-target pairs $(\bm{x}_t, \bm{y}_t)$.

\subsection{Generation Model}
\begin{figure}
\centering
\def\svgwidth{\columnwidth}
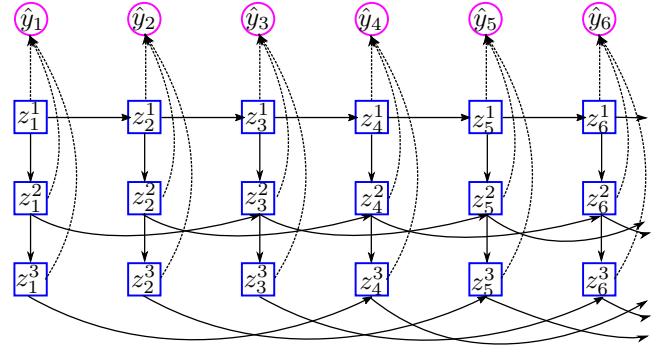
\caption{Generation Model. Solid line: transition neural network. Dashed line: emission neural network.}
\label{fig:gen}
\end{figure}

\begin{figure}[H]
\centering
\def\svgwidth{2.6in}
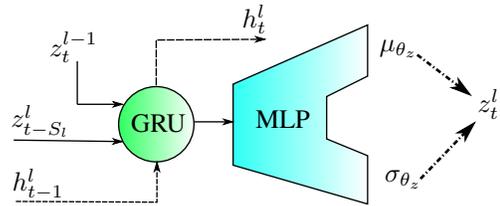
\caption{Transition Framework. Dashed lines show the transition of hidden states of GRU.}
\label{fig:trans}
\end{figure}

Figure \ref{fig:gen} shows the generation model of our MP-DMM. The generation process of our MP-DMM follows the transition and emission framework, which is obtained by applying multiple deep RNN to non-linear hierarchical state space models. Here the RNN is realized by GRU \cite{gru}. And latent variables in different layer have different transition steps. The generation model is carefully designed to incorporate the various state transitions and auxiliary connections in order to capture the multiple temporal dependencies present in sequential data. Here we denote all the weights in generation model as ${\theta}$. And the parameters for emission and transition framework are denoted as $\theta_y$ and $\theta_z$ respectively, i.e., $\theta=\{\theta_y, \theta_z\}$.
\\
{\bf Transition Framework} Specifically, we design the transition process of the latent state $\bm{z}$ to capture the hierarchical structure for multiple temporal dependencies with multiple transition steps. In each non-bottom layer $l$, at time step $t\ge1$, the next latent variable is sampled from a multi-variable Gaussian distribution,  conditioned on the last-period latent variable at this layer and the current latent variable at last layer. So,
\begin{equation}
\bm{z}_t^l\sim\mathcal{N}(\mu_{\theta_z}(\bm{z}_t^{l-1}, \bm{z}_{t-S_l}^l), \sigma_{\theta_z}(\bm{z}_t^{l-1}, \bm{z}_{t-S_l}^l)) \label{gen_trans}
\end{equation}
where $S_l$ is the transition period at layer $l$, the mean function $\mu_{\theta_z}$ and diagonal variance $\sigma_{\theta_z}$ come from outputs of a GRU further processed by a multi-layer perceptron (MLP). Here the GRU is to capture the temporal dependencies in each layer. The transition model for layer $l\ge1$ is shown in the Figure \ref{fig:trans}. 
The latent variable at the first layer is only conditioned on the last latent variable at same layer. We assume the prior of all latent variables at time $t\le0$ follow standard multi-variable Gaussian distribution, i.e., $\bm{z}_t^{1:L}\sim\mathcal{N}(0, \mathbb{I})$ for $t\le0$.
\\
{\bf Emission Framework} At certain time step $t\ge1$, the output $\hat{y}_t$ is sampled from a multi-variable Gaussian distribution conditioned on all the latent variables at time $t$. Specifically, assuming hierarchical latent variables have $L$ layers, then it is
\begin{equation}
\bm{\hat{y}}\sim\mathcal{N}(\mu_{\theta_y}(\bm{z}_t^{0},\ldots, \bm{z}_t^{L-1}), \sigma_{\theta_y}(\bm{z}_t^{0},\ldots, \bm{z}_t^{L-1})) \label{gen_emission}
\end{equation}
where $\hat{\bm{y}}_t$ is output at time $t$. Based on the model above, we can factorize the likelihood of generation model as below
\begin{eqnarray}
\lefteqn{p_{\theta}\big(\bm{y}_{1:T}, \bm{z}_{1:T}^{1:L}|\bm{z}_{-S_L+1:0}^{1:L}\big)} \nonumber \\
&=&p_{\theta}\big(\bm{y}_{1:T}|\bm{z}_{1:T}^{1:L}\big)p_{\theta}\big(\bm{z}_{1:T}^{1:L}|\bm{z}_{-S_L+1:0}^{1:L}\big) \nonumber \\
&=&\prod_{t=1}^T p_{\theta}\big(\bm{y}_t|\bm{z}_t^{1:L}\big)\cdot\prod_{t=1}^T p_{\theta}\big(\bm{z}_t^{1:L}|\bm{z}_{t-S_L:t-1}^{1:L}\big) \nonumber \\
&=&\prod_{t=1}^T\prod_{l=1}^L p_{\theta_y}\big(\bm{y}_t|\bm{z}_t^{1:L}\big)\prod_{t=1}^T p_{\theta_z}(\bm{z}_t^1|\bm{z}_{t-1}^1) \nonumber  \\
&&\cdot \prod_{l=2}^L p_{\theta_z}\big(\bm{z}_t^l|\bm{z}_{t-S_l}^l, \bm{z}_t^{l-1}\big) \label{gen_fact}
\end{eqnarray}
where $S_L$ is the maximum transition step in the model, $\theta_y$ and $\theta_z$ are parameters for emission and transition frameworks respectively. So $\theta=\{\theta_y, \theta_z\}$.

\subsection{Inference Model}
Here we design inference model to mimic the structure of generation model. That's because the goal of variational inference is to use variational distribution formed by inference model to approximate the posterior distribution of latent variables in the generation model \cite{vi1}. The inference model should be easy to optimize and generate samples of latent variables. Assume the parameters of inference model are denoted as $\phi$. Based on the standard definition of variational inference \cite{vi1}, we build a variational distribution of latent variables $q_{\phi}$ based on inference model, and maximize the evidence lower bound (ELBO) $\mathcal{F}(\theta, \phi)\le\mathcal{L}(\theta)$ with respect to both $\theta$ and $\phi$. Following the standard derivation of ELBO, our optimization target can be formulated as
\begin{eqnarray}
\lefteqn{\mathcal{F}(\theta,\phi)=\mathbb{E}_{q_{\phi}}\big[\log p_{\theta}(\bm{y}_{1:T}|\bm{z}^{1:L}_{-S_L+1:T})\big]} \nonumber \\
&&-D_{\text{KL}}\big(q_{\phi}\big(\bm{z}^{1:L}_{1:T}|\bm{x}_{1:T}, \bm{z}^{1:L}_{-S_L+1:0}\big)\|p_{\theta}\big(\bm{z}^{1:L}_{1:T}|\bm{z}^{1:L}_{-S_L+1:0}\big)\big) \label{ELBO1}
\end{eqnarray} 
where the first expectation is under distribution $q_{\phi}(\bm{z}^{1:L}_{1:T}|$ $\bm{x}_{1:T}, \bm{z}^{1:L}_{0})$.

\begin{figure}
\centering
\def\svgwidth{\columnwidth}
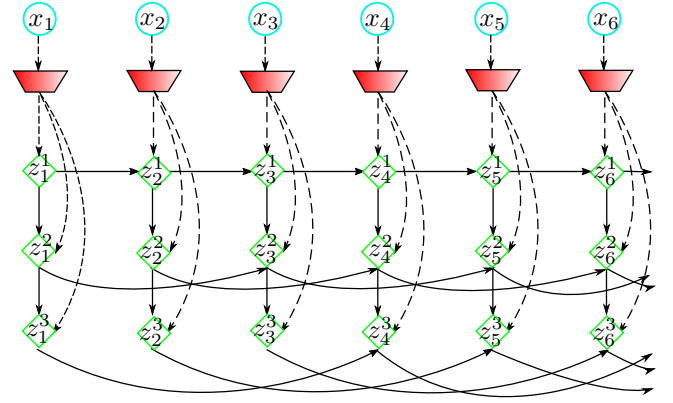
\caption{Inference Model. Solid line: transition neural network. Dashed line: input neural network. Red block: sparse input network $\varphi_{\phi_s}(\cdot)$.}
\label{fig:inf}
\end{figure}

Our inference model is shown in Figure \ref{fig:inf}. Generally, the variational distribution can approximate the posterior in three approaches, i.e., filtering, smoothing and bi-direction \cite{krishnan}. In this paper, we only construct the variational distribution in filtering setting, where the latent variables are dependent on input variables upto current time step. That's because the time series in practice are always processed in online approach, where the future values cannot be obtained in advance. 
\\
{\bf Transition Framework} Similar as generation model, at each time t, the latent variable in first layer, $\bm{z}_t^1$, is sampled from the multi-variable Gaussian conditioned on current input $\bm{x}_t$ and previous state $\bm{z}_{t-1}^1$. The latent variable in higher layer $l\ge1$, $\bm{z}_t^l$, is sampled from the multi-variable Gaussian conditioned on current input $\bm{x}_t$, the state in last period $\bm{z}_{t-S_l}^1$ and current state in last layer $\bm{z}_t^{l-1}$. The transition framework for latent variables in all layers is also realized by a GRU following a MLP, which produces the mean and diagonal for the distribution of next latent variables. 
\\
{\bf Input Network} In order to select most effective input variables, we add a sparse neural network $\varphi_{\phi_s}(\cdot)$ to pre-process the input observations, which is parameterized by $\phi_s$. This network is implemented by a sparse neural network with Horseshoe prior. So $\phi_s$ is treated as random variables. The details will be introduced in the next section.

Then, based on the discussion above and the Markov property of latent variables, we can factorize the variational distribution $q_{\phi}$ as below
\begin{eqnarray}
&&q_{\phi}\big(\bm{z}_{1:T}^{1:L}|\bm{x}_{1:T}, \bm{z}_{-S_L+1:0}^{1:L}\big)=\prod_{t=1}^T q_{\phi}\big(\bm{z}_t^{1:L}|\bm{z}_{t-S_L:t-1}^{1:L}, \bm{x}_{1:t}\big) \nonumber \\
&=&\prod_{t=1}^T q_{\phi_z}\big(\bm{z}_t^1|z_{t-1}^1, \varphi_{\phi_s}(\bm{x}_{t})\big) \nonumber \\
&&\cdot\prod_{l=2}^L q_{\phi_z}\big(\bm{z}_t^l|\bm{z}_{t-S_l}^l, \bm{z}_t^{l-1}, \varphi_{\phi_s}(\bm{x}_{t})\big) \label{inf_fact}
\end{eqnarray}
where each distribution $q$ is implemented as multi-variable Gaussian parameterized by $\phi_z$. Therefor $\phi=\{\phi_s, \phi_z\}$. 

\subsection{Sparse Input Network}
\begin{figure}[H]
\centering
\def\svgwidth{3in}
%% Accompanies image file 'bnn.pdf' (pdf, eps, ps)
\begingroup%
  \makeatletter%
  \providecommand\color[2][]{%
    \errmessage{(Inkscape) Color is used for the text in Inkscape, but the package 'color.sty' is not loaded}%
    \renewcommand\color[2][]{}%
  }%
  \providecommand\transparent[1]{%
    \errmessage{(Inkscape) Transparency is used (non-zero) for the text in Inkscape, but the package 'transparent.sty' is not loaded}%
    \renewcommand\transparent[1]{}%
  }%
  \providecommand\rotatebox[2]{#2}%
  \newcommand*\fsize{\dimexpr\f@size pt\relax}%
  \newcommand*\lineheight[1]{\fontsize{\fsize}{#1\fsize}\selectfont}%
  \ifx\svgwidth\undefined%
    \setlength{\unitlength}{286.08860026bp}%
    \ifx\svgscale\undefined%
      \relax%
    \else%
      \setlength{\unitlength}{\unitlength * \real{\svgscale}}%
    \fi%
  \else%
    \setlength{\unitlength}{\svgwidth}%
  \fi%
  \global\let\svgwidth\undefined%
  \global\let\svgscale\undefined%
  \makeatother%
  \begin{picture}(1,0.7256944)%
    \lineheight{1}%
    \setlength\tabcolsep{0pt}%
    \put(0,0){\includegraphics[width=\unitlength,page=1]{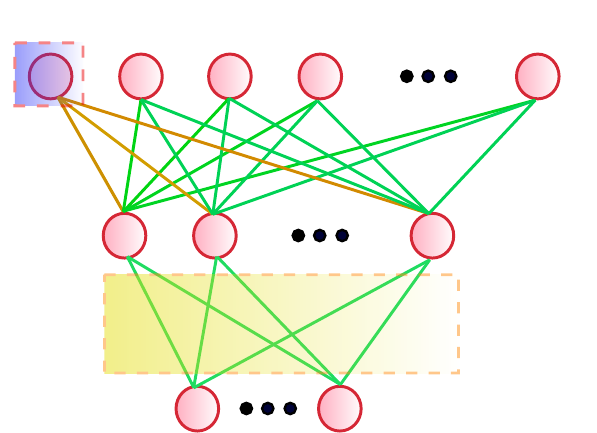}}%
    \put(0.79373093,0.16763464){\color[rgb]{0,0,0}\makebox(0,0)[lt]{\lineheight{1.25}\smash{\begin{tabular}[t]{l}$\nu_2$\end{tabular}}}}%
    \put(0.04624663,0.6858589){\color[rgb]{0,0,0}\makebox(0,0)[lt]{\lineheight{1.25}\smash{\begin{tabular}[t]{l}$\tau_{11}$\end{tabular}}}}%
  \end{picture}%
\endgroup%

\caption{Input Neural Network. Brown lines: weights $\phi_{s,kl}$, also controlled by vector $\beta_{11}$.}
\label{fig:bnn}
\end{figure}

The input neural network $\varphi_{\phi_s}(\cdot)$ is to select informative variables from input observations. In order to make the model simple and prevent overfitting, we make the input network to be 2-layer MLP. The number of neurons in the first layer, denoted as $N_1$, is equal to the input dimension. And the second layer has $N_2$ neurons. The output has dimension of $N_3$. We define $\phi_{s, kl}$ as the vector weights associated with the $k$-th neuron at $l$-th layer, i.e., $\phi_{s, k1}\in\mathbb{R}^{N_2}$ and $\beta_{2l}\in\mathbb{R}^{N_3}$. Since we formulate input network as Bayesian neural network, we have, $\beta_{kl}\sim\mathcal{N}(0,\mathbb{I})$, to represent the randomness of weight vector $\phi_{s,kl}$. And $\tau_{kl}$ and $\nu_{l}$ are scale parameters to control the local sparsity of $k$-th neuron of $l$-th layer, and the overall sparsity of $l$-th layer, respectively. According to \eqref{hs_prior2}, $\phi_{s,kl}=\tau_{kl}\nu_l\beta_{kl}$. The definitions of $\tau_{kl}, \nu_{l}$ and $\beta_{kl}$ is visualized in Figure \ref{fig:bnn}.

Although the prior in \eqref{hs_prior} is learnable in variational inference, it is difficult for regular exponential family variational approximation to capture the thick Cauchy tails, and variational distribution based on Cauchy family can lead to gradients with high variance. Following \cite{invgamma, invgamma2}, we adopt two hierarchical inverse Gamma priors for each scale parameter of sparsity, i.e.,
\begin{eqnarray}
\lefteqn{a\sim C^+(0,b)}\nonumber \\ 
&\Longleftrightarrow & a\sim\text{Inv-Gamma}(\frac{1}{2}, \frac{1}{\lambda}); \lambda\sim\text{Inv-Gamma}(\frac{1}{2}, \frac{1}{b^2}) \label{inv_gamma}
\end{eqnarray}
This prior applies for all $\tau_{kl}$ and $\nu_l$. 

However, this prior is still difficult to sample, and we need to propose a tractable variational distribution $q_s(\phi_s|\zeta)$ to approximate the true posterior of $\phi_{s,kl}$. And $q_s$ is paramaterized by $\zeta$. Define $\beta_l\in\mathbb{R}^{N_l\times N_{l+1}}$ as the matrix, each row of which is vector $\beta_{kl}$. Then following \cite{invgamma2, invgamma3}, we use fully factorized variational family for $\nu_l$ and $\tau_{kl}$, and structured variational family for $\beta_{l}$, shown as below,
\begin{eqnarray}
q_s(\phi_s|\zeta)&=&\prod_{l=1}^{L_s}q_s(\nu_l|\zeta_{\nu_l})\mathcal{M}\mathcal{N}(\beta_l|M_{\beta_l}, U_{\beta_l}, V_{\beta_l}) \nonumber \\
&&\cdot\prod_{k=1}^{N_l}q_s(\tau_{kl}|\zeta_{\tau_{kl}})\delta_{\phi_{s,kl}}(\tau_{kl}\nu_l\beta_{kl}) \label{q_s}
\end{eqnarray}
where $L_s$ is the number of layers in input network. Here the randomness of all weights in layer $l$, i.e. $\beta_l$, has the variational approximation formulated by matrix Normal distribution, which is to capture the dependencies among weights in the same layer. And the variational distribution for non-negative scale parameters $\tau_{kl}$ and $\nu_l$ follow log-Normal distributions, i.e., $q_s(\ln\tau_{kl}|\zeta_{\tau_{kl}})=\mathcal{N}(\mu_{\tau_{kl}}, \sigma_{\tau_{kl}})$ and $q_s(\ln\nu_{l}|\zeta_{\nu_{l}})=\mathcal{N}(\mu_{\nu_{l}}, \sigma_{\nu_{l}})$. For simpler computation and better regularization, the matrix $V_l$ is approximated by a diagonal matrix, and $U_l$ is approximated by $\Pi+u_l u_l^T$, where $u_l$ is a vector and $\Pi$ is a diagonal matrix.

\subsection{Learning the Parameters}
According to Bayesian variational inference, we jointly learn generation and inference parameters $\theta$ and $\phi$ by optimizing ELBO in \eqref{ELBO1}. However, due to factorization of generation and inference models in \eqref{gen_fact} and \eqref{inf_fact}, the ELBO can be reformulated as
\begin{eqnarray}
\lefteqn{\mathcal{F}(\theta, \phi_s, \phi_z)=\mathcal{F}(\theta, \phi)=\sum_{t=1}^T\mathbb{E}_{\tilde{q}_{\phi}(\bm{z}_{t}^{1:L})}\log p_{\theta_y}\big(\bm{y}_t|\bm{z}_{t}^{1:L}\big)} \nonumber \\	
&&-\sum_{t=1}^T\mathbb{E}_{\tilde{q}_{\phi}(\bm{z}_{t-1}^1)}D_{\text{KL}}\bigg(q_{\phi}\bigg(\bm{z}_t^1|\bm{x}_{t}, \bm{z}^1_{t-1}\bigg)\bigg\|p_{\theta}\bigg(\bm{z}_t^1|\bm{z}_{t-1}^1\bigg)\bigg) \nonumber \\
&&-\sum_{t=1}^T\sum_{l=2}^L\mathbb{E}_{\tilde{q}_{\phi}(\bm{z}_{t-S_l}^l, \bm{z}_t^{l-1})} \nonumber \\
&&D_{\text{KL}}\bigg(q_{\phi}\bigg(\bm{z}_t^l|\bm{x}_{t}, \bm{z}^l_{t-S_l}, \bm{z}_t^{l-1}\bigg)\bigg\|p_{\theta}\bigg(\bm{z}_t^l|\bm{z}_{t-S_l}^l, \bm{z}_t^{l-1}\bigg)\bigg) \label{elbo_fac}
\end{eqnarray}
where distribution $\tilde{q}_{\phi}$ is the marginal of all other latent variables and input observations. 

Since the input network parameters $\phi_s$ in $\phi$ are modeled as random variables, we need to introduce another ELBO to learn its optimal variational approximate distributions. Combining its prior \eqref{inv_gamma} and variational approximations $q_s$ in \eqref{q_s}, the overall ELBO should be dependent on $\zeta$, rather than $\phi_s$. Then, with the factorized ELBO above \eqref{elbo_fac}, the optimization objective should be,
\begin{eqnarray}
\lefteqn{\mathcal{E}(\theta, \phi_z, \zeta)=\mathbb{E}_{q_s(\cdot|\zeta)}\bigg[\mathcal{F}(\theta, \phi_s, \phi_z)\bigg|\phi_s\bigg]} \nonumber \\
&&+ \sum_{l=1}^{L_s}\mathbb{E}_{q_s(\cdot|\zeta)}\bigg[\log\text{Inv-Gamma}\bigg(\nu_l\bigg|\frac{1}{2}, \frac{1}{\upsilon_l}\bigg) \nonumber \\
&&+ \log\text{Inv-Gamma}\bigg(\upsilon_l\bigg|\frac{1}{2}, \frac{1}{b_g^2}\bigg)\bigg] \nonumber \\
&& + \sum_{k=1}^{N_l} \mathbb{E}_{q_s(\cdot|\zeta)}\bigg[\log\text{Inv-Gamma}\bigg(\tau_{kl}\bigg|\frac{1}{2}, \frac{1}{\lambda_{kl}}\bigg) \nonumber \\
&&+ \log\text{Inv-Gamma}\bigg(\lambda_{kl}\bigg|\frac{1}{2}, \frac{1}{b_0^2}\bigg)\bigg] \nonumber \\
&& + \mathbb{E}_{q_s(\cdot|\zeta)}\bigg[\log \mathcal{N}(\beta_l|0, \mathbb{I})\bigg] + \mathcal{H}(q_s(\cdot|\zeta)) \label{obj}
\end{eqnarray}
where $b_0$ and $b_g$ are hyperparamters, and $\mathcal{H}$ is the entropy term. There are closed form expressions for cross-entropy between log-Normal and Inv-Gamma, and entropies for log-Normal and matrix Normal distributions already have standard formulas to use. The first term with expectation is difficult to compute. Here we adopt stochastic gradient descent and black box vairational inference \cite{vi2, horseshoe} to evaluate it. It is to approximate the expectation by a Monte Carlo approach with reparameterization trick \cite{vae}, and we can have unbiased estimate of gradients. Then the objective \eqref{obj} can be optimized by ADAM \cite{adam}.

\section{Experiment Results}
In experiments, we evaluate the proposed algorithm on two datasets. The first dataset is generated by pre-defined model, and we evaluate the negative log likelihood (NLL) on this dataset. The second dataset comes from Rossmann sales data, open to the public on Kaggle platform, and we perform sales prediction in this case. In both experiments, we compare proposed algorithm with vanila LSTM \cite{lstm}, VRNN \cite{vrnn} and SRNN \cite{srnn}, which are popular sequential data models proposed recently. And inference models in both VRNN and SRNN don't have backward RNN, since every model should be working on filtering setting.
\subsection{Synthesis Dataset}
Assume the target data $\bm{y}_t\in\mathbb{R}^5$ is generated as below \cite{boning}
\begin{equation}
\bm{y}_t=\bm{\mu}_t+\bm{\delta}_t+\bm{X}_t\bm{\beta}+\bm{\epsilon}_t \nonumber
\end{equation}
where $\bm{\mu}_t, \bm{\delta}_t, \bm{\epsilon}\in\mathbb{R}^5$, and $\bm{X}_t\in\mathbb{R}^{5\times50}$ is the input observations at time $t$, and $\bm{\beta}\in\mathbb{R}^{50}$ is a sparse coefficient vector of input data. Each target variable has 10 input variables, which are generated by an AR(1) process with coefficient $0.6$ and standard error $0.1$. In $\bm{X}_t$, $i$-th row contains input observations for $i$-th target variable at time $t$, where elements from $10i$-th to $10(i+1)-1$-th positions are non-zeros while others are set to zeros. Each data sequence has the length of $100$.

 Here we assume that for each target variable, there are 10 input variables, and only first three input variables are related with target data, i.e., the last 7 elements in $\bm{\beta}$ are zeros. For $i=1,\ldots,5$, the trend is generated from $\mu_{it}\sim\mathcal{N}(0.8\mu_{i,t-1}, 0.2^2)$ with $\mu_{i0}=1$. And the inherent period of $\bm{\delta}_t$ is 7, then for $i=1,\ldots,5$,
\begin{equation}
\delta_{it}=\alpha\times\cos(2\pi t/7) + \alpha\times\sin(2\pi t/7) \nonumber
\end{equation} 
where $\alpha$ is set to different values in experiments to generate different datasets. Moreover, the observation error $\bm{\epsilon}_t$ is sampled from the multivariate Normal distribution $\mathcal{N}(0, \bm{\Sigma})$ with precision given as below,
\begin{equation}
\bm{\Sigma}^{-1}=\begin{bmatrix}
   10 & 5 & 0 & 0 & 0 \\
   5 & 10 & 5 & 0 & 0 \\
   0 & 5 & 10 & 5 & 0 \\
   0 & 0 & 5 & 10 & 5 \\ 
   0 & 0 & 0 & 5 & 10 
  \end{bmatrix}  \nonumber
\end{equation}
Every evaluated model has two layers of latent variables, and in the proposed model, the second layer has transition period of 7, i.e., $S_1=1$ and $S_2=7$. In experiment, the coefficient $\alpha$ in trends is set to be $0.1, 0.15$ and $0.20$. The performance comparison is shown in Table \ref{exp1}.  

\begin{table}[H]
\centering
\begin{tabular}{c|c|c|c|c}
     & Proposed & VRNN & SRNN & LSTM \\\hline \hline
$\alpha=0.1$ & \bf -10.325 & -11.811 & -12.987 & -15.645 \\\hline
$\alpha=0.15$ & \bf -10.013 & -12.156 & -13.102 & -16.860 \\\hline
$\alpha=0.2$ & \bf -9.550 & -12.684 & -13.523 & -17.543
\end{tabular}
\caption{Negative Log-likelihood Comparison.}
\label{exp1}
\end{table}

\subsection{Real-world Dataset}
In the second experiment, we perform one-step prediction on sales data from Rossmann. Besides sales records, it contains custom flow information, holiday information, store type, assortment and promotion activities in the same time interval. Moreover, we added the oil price, EURO to USD currency exchange ratio history, and DAXI (German stock market index) history into input variables, which are all publicly available. Most of them may be irrelevant to sales prediction, which is test our algorithm can avoid such influence or not. The input and target data corresponding to dates when stores are closed will be omitted in the experiment. Based on data inspection, the sales data has three inherent periods, daily, weekly, and monthly periods. So, the models in comparison all have three layer of latent variables, and in the proposed model, the transition period in every layer will be $S_1=1, S_2=7$ and $S_3=30$. 

Different from data modeling, the prediction aims at accuracy, rather than higher likelihood. Based on experimental experience, the prediction can be improved when the KL divergence term in \eqref{ELBO1} is multiplied by a coefficient $0<\xi<1$, i.e., 
\begin{eqnarray}
\lefteqn{\mathcal{F}_{\xi}(\theta,\phi)=\mathbb{E}_{q_{\phi}}\big[\log p_{\theta}(\bm{y}_{1:T}|\bm{z}^{1:L}_{-S_L+1:T})\big]} \nonumber \\
&&-\xi D_{\text{KL}}\big(q_{\phi}\big(\bm{z}^{1:L}_{1:T}|\bm{x}_{1:T}, \bm{z}^{1:L}_{-S_L+1:0}\big)\|p_{\theta}\big(\bm{z}^{1:L}_{1:T}|\bm{z}^{1:L}_{-S_L+1:0}\big)\big) \nonumber
\end{eqnarray} 
Then the learning objective \eqref{obj} also incorporates $\xi$. All the algorithms are evaluated at different choices of $\xi$ in experiments. The performance metric is Root Mean Square Percentage Error (RMSPE), i.e.,
\begin{equation}
\text{RMSPE}=\sqrt{\frac{1}{T}\sum_{t=1}^T\frac{\|\bm{y}_t-\hat{\bm{y}}_t\|^2_2}{\|\bm{y}_t\|^2_2}} \nonumber
\end{equation}
where $\bm{y}_t$ and $\hat{y}_t$ are target data and the prediction result respectively. Then the performance comparison is shown in Table \ref{exp2}.

\begin{table}[H]
\centering
\begin{tabular}{c|c|c|c|c}
     & Proposed & VRNN & SRNN & LSTM \\\hline \hline
$\xi=1e-2$ & 0.121 & \bf 0.115 & 0.125 & 0.203 \\\hline
$\xi=1e-3$ & \bf 0.110 & 0.119 & 0.114 & 0.182 \\\hline
$\xi=1e-4$ & \bf 0.119 & 0.122 & 0.120 & 0.173
\end{tabular}
\caption{RMSPE Comparison.}
\label{exp2}
\end{table}

\section{Conclusion}
In this work, we proposed a new model for multi-period time series modeling, with input variable selection implemented by Bayesian sparse learning. The multi-period time series are commonly seen in real life, such as the sales data are governed by daily trend, weekly trend and monthly trends. Meteorological data and oceanic data are also in this case. We build a hierarchical latent variable model, where different layers have different transition steps, matching the inherent periods governing the dynamics of time series data. In order to discard uninformative input observations, we also introduce a sparse Bayesian neural network, following Horseshoe prior with non-centered parameterization. The performance is evaluated on both synthesis data and real-world data.

\end{document}